# A Dynamic Modelling Framework for Human Hand Gesture Task Recognition

**Sara Masoud[1], Bijoy Chowdhury[1], Young-Jun Son[1], Chieri Kubota[2], Russell Tronstad[3]**

[1]Department of Systems and Industrial Engineering, University of Arizona

[2]Department of Horticulture and Crop Science, Ohio State University

[3]Department of Agriculture and Resource Economics, University of Arizona

[1,3]Tucson, Arizona, USA

[2]Columbus, Ohio, USA

## Abstract

Gesture recognition and hand motion tracking are important tasks in advanced gesture based interaction systems. In this paper, we propose to apply a sliding windows filtering approach to sample the incoming streams of data from data gloves and a decision tree model to recognize the gestures in real time for a manual grafting operation of a vegetable seedling propagation facility. The sequence of these recognized gestures defines the tasks that are taking place, which helps to evaluate individuals' performances and to identify any bottlenecks in real time. In this work, two pairs of data gloves are utilized, which reports the location of the fingers, hands, and wrists wirelessly (i.e., via Bluetooth). To evaluate the performance of the proposed framework, a preliminary experiment was conducted in multiple lab settings of tomato grafting operations, where multiple subjects wear the data gloves while performing different tasks. Our results show an accuracy of 91% on average, in terms of gesture recognition in real time by employing our proposed framework.

## 1. Introduction

Gesture recognition technology is defined to automatically analyse the changes of shape or gesture of the human body parts to determine a user's intentions in order to deliver a proper response [1]. Gesture recognition has gained popularity in recent years due to its ability to connect humans and machines in terms of communication, interactions, and control. Tracking different body parts such as arms, hands, and face to recognize patterns have been studied in the literature. In addition, hand gesture is commonly used in most of human computer interaction interfaces including both static and dynamic gesture recognition [2, 3]. These gesture recognitions take place by utilizing techniques that can be categorized into two main groups: 1) the vision based approaches and 2) wearable sensor based approaches [4].

Vision-based approaches usually use sequences of images from video sensor for recognition [5]. Generally, they can be categorized into appearance based techniques and 3D model based techniques [6]. Appearance based techniques rely on the difference between the parameters of the modelled limb under-study based on the features selected from the images and features of the input video, while 3D model based techniques focus on defining those parameters by comparing the 3D input image and possible 2D appearance projected by the 3D model considering all degree of freedoms of that limb [6]. In general, appearance-based techniques provide faster recognition, while 3D hand model-based techniques offer higher accuracy [6]. Although vision-based methods excel in perfect indoor activities, it has numerous restrictions such as space, interruption of the light, and interference by environment [1, 7].

On the other hand, the wearable sensor based motion recognition employs the retrieval of continuous signals, such as acceleration from accelerometer for recognition. This method uses sensors to digitize motions into multi-parametric data [8]. The wearable sensor based motion recognition provides various advantages such as freedom of occlusion, high performance in a complex environment, and more detailed coverage [3]. Data gloves are a wearable sensing technology that can help recognise tasks that are taking place by hands [9]. These gloves can include multiple motion sensors such as accelerometer, gyroscope, bend sensor, and force sensor [1]. Data gloves are widely utilized for training or control in many different applications such as virtual reality, robotics, biomechanics, and surgery [10]. For example, [11] has introduced a cost-efficient data glove for training in virtual surgeries. [7] has presented a data glove with inertial and magnetic sensors for controlling robotic arms. [12] has utilized a data glove with embedded inertial sensors to help with evaluation of hand functions of stroke patients.

Although lots of research has taken place in the field of gesture recognition via data gloves, we are proposing a framework which can track the tasks in real time with a high level of accuracy. In this study, we are using custom made VMG 30 data gloves which include 10 embedded bend sensors, 4 abduction sensors, 1 palm arch sensor, and 1 thumb cross over sensor in each glove. The customization of data gloves includes the removal of finger tips which alongside with wireless (i.e., via Bluetooth) communication, reduces the hindrance of wearing the gloves for the subjects. Figure 1 displays the kinematic based graphs, based on which the glove captures the motions of the hand.

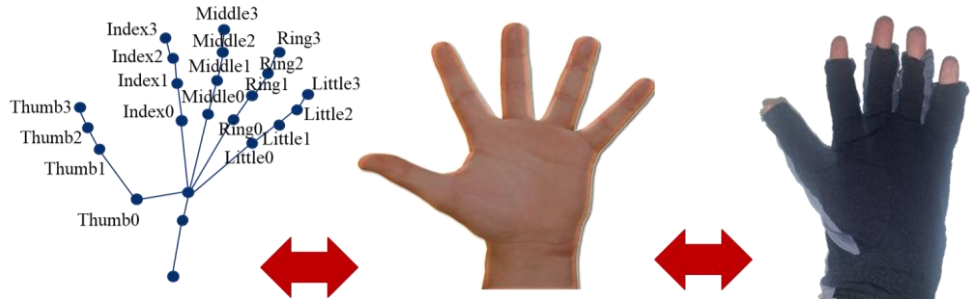

Figure 1: Kinematic based hand model for hand motion tracking

While the focus of this work is on the methodological side, to illustrate the efficiency of the proposed framework we have conducted some preliminary experiments, where we have studied a tomato grafting operation. Although grafting is a horticultural technique that improves crop yields as well as disease resistance, it could not become a commercial practice until the twentieth century due to high production cost, which is a result of intensive labor requirements [7, 13, 14]. However, technologies similar to what is proposed in this work can help to reduce these labor costs by evaluating workers' performance and improving labor efficiency of the grafting industry.

## 2. Methodology

The overview of the proposed hand gesture recognition framework is described in Figure 2. The framework consists of two main phases: which are 1) training and 2) prediction. The prediction phase follows the training phase continuously with a time gap.

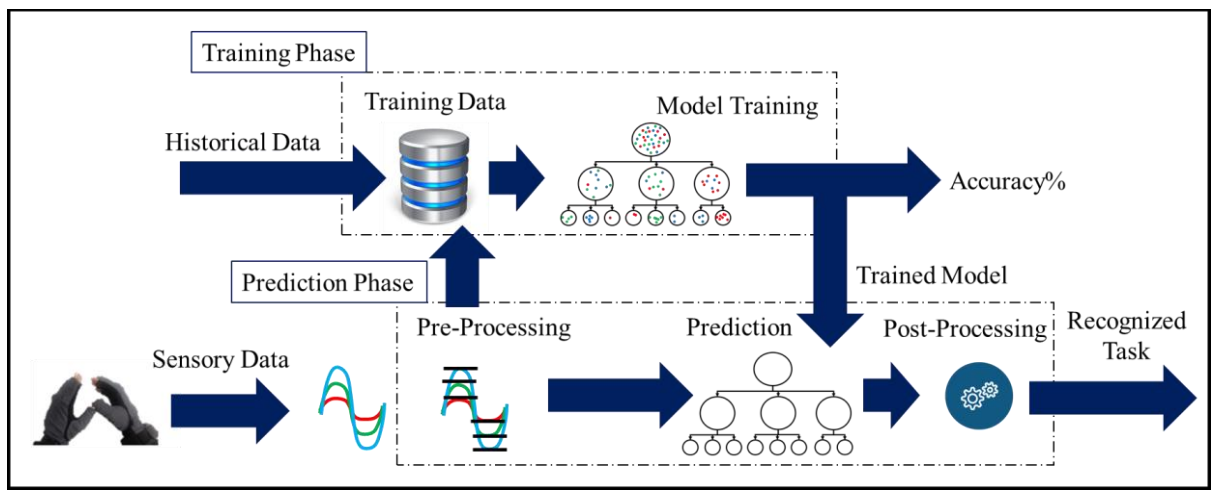

Figure 2: The proposed task recognition framework

Starting with the training phase, this phase uses the 70% of the historical pre-processed labeled dataset to train a classification model and the other 30% to obtain the accuracy level of the mentioned model. As soon as the training for the processed historical data are finished, the prediction phase starts to predict the task by utilizing the trained model. Afterwards, these two phases run in parallel in order to let the training phase keep the trained model up-to-date and the prediction phase to recognize the tasks. The classification algorithm that we are using in this framework is the decision tree (i.e., C4.5) [15]. We train the model based on pre-processed historical data. The same model is used in the prediction phase and for each of the data streams, the framework predicts the tasks which has generated that stream. A critical point in this framework is the continuous updating of training dataset based the streams of sensory data. As mentioned before, each decision tree is a classification algorithm and is utilized for recognizing the tasks. As a result, a labeled up-to-date training dataset is a key requirement for accurate prediction. To achieve this, we perform pre-processing before adding the sensory data to the dataset of training data. Figure 3 illustrates this pre-processing step.

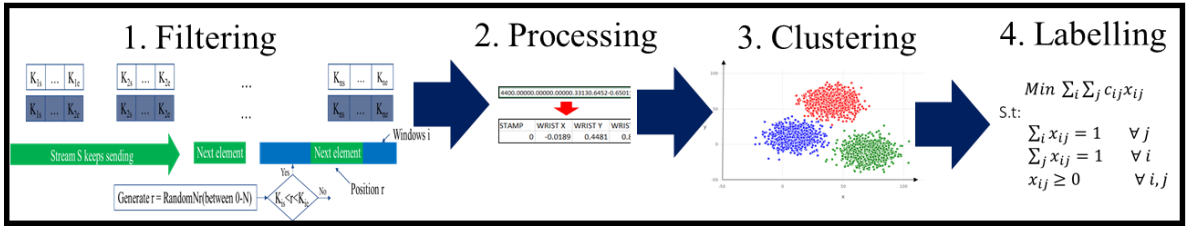

Figure 3: Illustration of the pre-processing step

As shown in Figure 3, the first step in pre-processing is filtering to enable us to track the tasks in real time. Here, we use the sliding window filtering technique which includes a set of 1D arrays with equal sizes and equal gaps between arrays. For each stream of data, we generate a random number. If the random number falls within one of the arrays, we update the array by that stream of data, otherwise we let the data stream pass through. In addition, to make sure that we have a sample that can realistically represent the system in real time we have defined time constraints in arrays as well. So, if an array is not updated within a time interval, that array will be updated with the next incoming data stream. In the next step, we structuralize the data streams into its basic elements which are the time stamp and the coordination of the nodes as shown in Figure 1. Next, we need to label this dataset. Given the fact that the number of the tasks to be recognized is a known number, we utilize the k-means [16] clustering algorithm to define clusters based on the number of tasks. For labeling these defined clusters, we use an assignment problem [17] to assign the appropriate labels to their related clusters in order to minimize the summation of errors (i.e., $e_{ij}$). These errors are defined based on the difference between the averages of each cluster of the training dataset which are labeled (i.e., $i \in I$) and the averages of each sensory clusters (i.e., $j \in J$). Set $I$ (i.e., the labels) contains Scion Cutting, Rootstock Cutting, Rootstock Clipping, and Joining which are the main tasks in grafting.

$$\min \sum_{i \in I} \sum_{j \in J} e_{ij} x_{ij} \quad (1)$$

Subject to:

$$\sum_{i \in I} x_{ij} = 1 \quad \forall j \in J \quad (2)$$

$$\sum_{j \in J} x_{ij} = 1 \quad \forall i \in I \quad (3)$$

$$x_{ij} \geq 0 \quad \forall i, j : i \in I, j \in J \quad (4)$$

The objective (i.e., Equation 1) is assigning the labels to the clusters in order to minimize the summation of the errors. Equation 2 ensures that each label is exactly assigned to one cluster, while Equation 3 guarantees that for each cluster, a label is assigned. Equation 4 constrains the physical boundaries on the decision variable $x_{ij}$ which define whether label $i$ is assigned to cluster $j$.

The last step in the prediction phase is the post-processing. In this step, we define characteristics of the recognized tasks within each grafting cycle such as starting time, duration (i.e., processing time), and ending time of the recognized tasks in order to evaluate the workers performances over time to detect abnormalities or improvement opportunities such as bottlenecks. These processing times are defined as the difference between the detection time of the sequential tasks as shown in Figure 4.

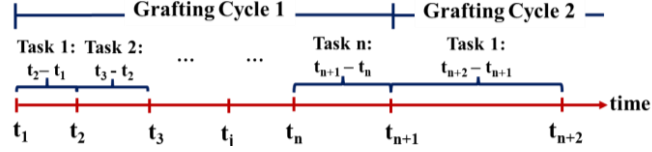

Figure 4: Processing time estimation based on sequential task recognition

We have employed the Bayesian approach [18] to generate advanced analysis in order to detect abnormalities. By incorporating prior information regarding the processing time (i.e., $p(x_k|z_{1:k-1})$) as provided in the historical dataset for each task, and given the streams of sensory data (i.e., $z_k$), we update a 99% confidence interval around the posterior mean of processing time (i.e., $p(x_k|z_{1:k})$). Any mean of processing time distribution that is not within this confidence interval boundaries, is labelled as an abnormality. Equation 5 displays the relation between the prior and posterior distributions in the Bayesian approach.

$$p(x_k | z_{1:k}) = \frac{p(z_k | x_k) p(x_k | z_{1:k-1})}{\int p(z_k | x_k) p(x_k | z_{1:k-1}) dx_k} \tag{5}$$

## 3. Experiment and Results

Tomato grafting operations consist of four major tasks: 1) cutting the scion seedling at a pre-defined angle (i.e., scion cutting), 2) cutting the rootstock seedling at the same angle as scion (i.e., rootstock cutting), 3) clipping the end cut of rootstock seedling (i.e., rootstock clipping) and 4) firmly joining the scion and rootstock together (i.e., joining). In this preliminary experiment, subjects follow the mentioned instructions. He/ she starts by cutting the scion and rootstock seedlings, clipping the end cut of rootstocks, and finally, joins the end cuts of scions and rootstocks firmly to each other.

For this experiment, a total of 110,000 data points has been employed as the historical training dataset. This dataset has been defined by asking three different subjects to perform grafting operations in lab settings while wearing the data gloves. In addition, this dataset has been updated by utilizing the incoming data streams. Figure 5 illustrates a snapshot of the initializing decision tree model trained by the historical dataset.

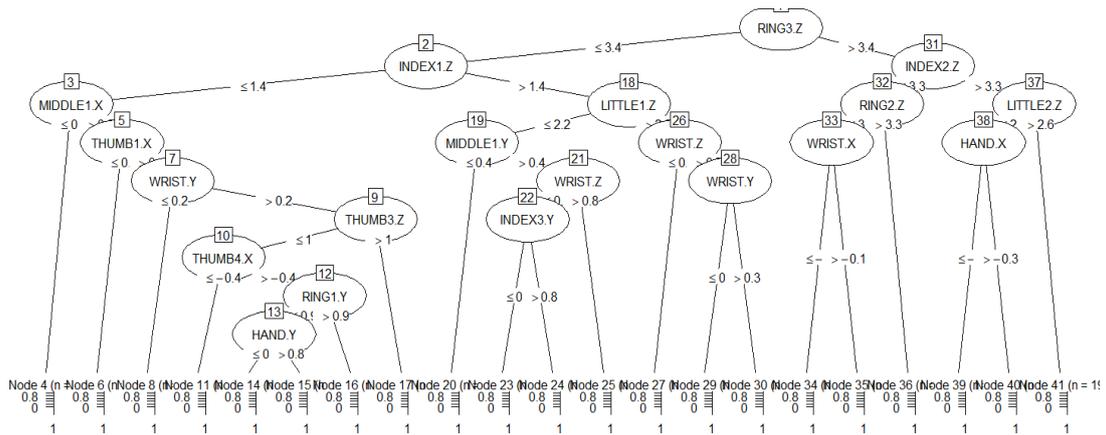

Figure 5: The first developed decision tree model

The models utilized in this study have an average size of 22 (i.e., the quantity of ending branches) and detection time of 0.09 seconds. The top 5 attributes in these decision trees are RING3.Z, INDEX1.Z, INDEX2.Z, MIDDLE1.X, and LITTLE2.Z which have the average contribution rate of 97.2%, 55.4%, 44.2%, 27.6%, and 25.7%, respectively.

As mentioned in Section 2, a major step in pre-processing is to label the incoming data streams. By employing the k-means algorithm, we define the four classes to cover the main tasks (i.e., scion cutting, rootstock cutting, rootstock clipping, and joining). For the top 5 attributers of the decision tree model, we defined the center of each class as shown in Table 1. These are the values which are used in the assignment model to define the error values (i.e., $e_{ij}$) in order to label the classes formed by the sampled data streams.

Table 1: Processing time estimation based on sequential task recognition

|  | Average | | | |
| --- | --- | --- | --- | --- |
|  | Scion Cutting | Rootstock Cutting | Rootstock Clipping | Joining |
| RING3.Z | 1.167 | 3.293 | 3.612 | 2.800 |
| INDEX1.Z | 0.430 | 2.059 | 2.400 | 2.022 |
| INDEX2.Z | 0.634 | 2.816 | 3.400 | 2.822 |
| MIDDLE1.X | -2.062 | 0.480 | -0.355 | -0.820 |
| LITTLE2.Z | 1.093 | 2.723 | 2.805 | 2.133 |
| Error | 5.85% | | | |

As illustrated in Table 1, an average error rate of 5.85% occurs in clustering the incoming data streams. Given this value, an average accuracy level of 91% has been achieved by utilizing the proposed framework. Table 2 displays a set of results.

Table 2: Illustration of results

| ID | Time Stamp | Detected Task | Processing Time (s) | Acceptance Range | Detected Abnormality |
| --- | --- | --- | --- | --- | --- |
| 117 | 2320 | Scion Cutting | 4.19 | [4.18, 4.20] | No |
| 118 | 2340 | Scion Cutting | | | |
| 137 | 2720 | Scion Cutting | | | |
| 138 | 2740 | Rootstock Cutting | 2.69 | [2.65, 2.72] | No |
| 139 | 2760 | Rootstock Cutting | | | |
| 151 | 3000 | Rootstock Cutting | | | |
| 152 | 3020 | Rootstock Clipping | 3.39 | [3.37, 3.45] | No |
| 168 | 3340 | Rootstock Clipping | | | |
| 169 | 3360 | Joining | 12.99 | [12.11, 13.01] | No |
| 170 | 3380 | Joining | | | |
| 233 | 4640 | Joining | | | |
| 234 | 4660 | Rootstock Clipping | 0.19 | [2,99 3.19] | Yes |
| 235 | 4680 | Scion Cutting | 4.39 | [4.23, 4.26] | No |
| 245 | 4700 | Scion Cutting | | | |
| 256 | 5100 | Scion Cutting | | | |
| 257 | 5120 | Rootstock Cutting | 4.75 | [3.94, 4.33] | Yes |
| 258 | 5140 | Rootstock Cutting | | | |
| 280 | 5520 | Rootstock Cutting | | | |
| 281 | 5600 | Rootstock Clipping | NA | NA | NA |

Table 2 illustrates the outcome of the proposed framework. The first column shows the ID of each data stream while the second column displays the time stamp related to each stream. The detected task associated with each task is placed in column 3. Forth column shows the processing time of each task. The Bayesian acceptance intervals are displayed in the sixth column. The last column describes whether any abnormalities have been detected.

## 4.  Conclusion and Future Work

In this paper, we have presented a hand gesture task recognition framework. This framework utilizes decision trees for classification purposes in real time. In addition, a pre-processing step including sliding windows filtering technique, k-means clustering algorithm, and assignment problem are utilized in order to present a real time labelled training set. To evaluate our proposed algorithm, a pilot experiment via VMG30 data gloves has been conducted to provide streams of hand gesture of a tomato grafting operation in a lab setting. Preliminary results have shown promising task recognition and abnormality detection with an accuracy level of 91%.

Although the proposed framework is generic and applicable to other domains, it needs to be specially trained for any new application with hand-crafted, structured datasets. While this research focuses on the conceptual framework of hand gesture-based task recognition, a more comprehensive experiment involving more grafting workers in a real vegetable grafting propagation facility is left for future research.

## Acknowledgements

This work has been supported by U.S. Department of Agriculture (USDA) – National institute of food and agriculture under project number 2016-51181-25404.